# Partial Label Learning with Focal Loss for Sea Ice Classification Based on Ice Charts

Behzad Vahedi, Benjamin Lucas, Farnoush Banaei-Kashani, *Professional Member, IEEE*, Andrew P. Barrett, Walter N. Meier, Siri Jodha Khalsa, *Senior Member, IEEE*, Morteza Karimzadeh, *Member, IEEE*

*Abstract*—Sea ice, crucial to the Arctic and Earth's climate, requires consistent monitoring and high-resolution mapping. Manual sea ice mapping, however, is time-consuming and subjective, prompting the need for automated deep learning-based classification approaches. However, training these algorithms is challenging because expert-generated ice charts, commonly used as training data, do not map single ice types but instead map polygons with multiple ice types. Moreover, the distribution of various ice types in these charts is frequently imbalanced, resulting in a performance bias towards the dominant class. In this paper, we present a novel GeoAI approach to training sea ice classification by formalizing it as a partial label learning task with explicit confidence scores to address multiple labels and class imbalance. We treat the polygon-level labels as candidate partial labels, assign the corresponding ice concentrations as confidence scores to each candidate label, and integrate them with focal loss to train a Convolutional Neural Network (CNN). Our proposed approach leads to enhanced performance for sea ice classification in Sentinel-1 dual-polarized SAR images, improving classification accuracy (from 87% to 92%) and weighted average F-1 score (from 90% to 93%) compared to the conventional training approach of using one-hot encoded labels and Categorical Cross-Entropy loss. It also improves the F-1 score in 4 out of the 6 sea ice classes.

Manuscript received xx; revised xx; accepted xx. Date of publication xx; date of current version xx. This work was supported by the U.S. National Science Foundation under Grants No. 2026962 and 2026865 (*Corresponding author: Morteza Karimzadeh*).

Behzad Vahedi and Morteza Karimzadeh are with the Department of Geography, University of Colorado Boulder, Boulder, CO, USA (e-mail: behzad@colorado.edu; Karimzadeh@colorado.edu).

Benjamin Lucas was with the Department of Geography, University of Colorado Boulder. He is now with the Department of Mathematics and Statistics, Northern Arizona University (e-mail: ben.lucas@nau.edu).

Farnoush Banaei-Kashani is with the Department of Computer Science, University of Colorado Denver, Denver, CO, USA (email: farnoush.banaei-kashani@ucdenver.edu).

Andrew P. Barrett, Walter N. Meier, and Siri Jodha Khalsa are with the National Snow and Ice Data Center (NSIDC), CIRES, University of Colorado Boulder, Boulder, CO, USA (e-mail: Andrew.Barrett@colorado.edu; walt@colorado.edu; khalsa@colorado.edu).

## 1. Introduction

Sea ice is a layer of frozen seawater that forms when the temperature of the surface of the water reaches its freezing point. Due to the low temperatures required for this process to occur, it is usually found in polar and sub-polar oceans and seas, though in winter it can extend equator-ward to 40–50° latitudes [1]. Arctic sea ice plays a crucial role in regulating global climate. It reflects solar radiation, helping maintain Earth's heat balance. Changes in ice coverage contribute significantly to global warming via the ice-albedo feedback loop. Furthermore, sea ice acts as a barrier, modulating heat, moisture, and gas exchanges between the ocean and the atmosphere. Therefore, Arctic sea ice changes have profound implications for ecosystems, atmospheric circulation, and global weather patterns [2]. Monitoring sea ice conditions and mapping its properties, such as type, extent, and concentration, are important for climate monitoring as well as marine navigation and near- or off-shore operations [1].

Ice packs with older Stages of Development (SoDs) are thicker and pose a greater hazard risk for marine navigation (based on the type of vessel) [3]. Thus, high-quality, high-resolution, and timely sea ice type maps play a crucial role in ensuring safe marine navigation and in reducing the environmental footprint (and risk) of vessels [4]. Ideally, navigators (vessels) require ice charts at high temporal and spatial resolutions (300 m or higher) to navigate safely [5].

The need for high-resolution and scalable mapping has led to several efforts to automate sea ice mapping. This, however, is a challenging task due to (a) the dynamic nature of sea ice, (b) ambiguous and similar signatures for different ice types (particularly in Synthetic Aperture Radar (SAR) imagery), (c) the effects of wind and weather on remotely sensed products, and (d) the effects of surface roughness and volume scattering on ice emissivity [6]–[10]. Therefore, sea ice charting is still primarily performed manually by sea ice analysts at national organizations such as the United States National Ice Center (USNIC), the Canadian Ice Service (CIS), and the Norwegian Meteorological Institute (MET Norway) by visually analyzing different data sources, including remotely-sensed optical, passive microwave, and SAR images; climatological model outputs; and in-situ measurements [6].

The increasing demand for sea ice products, the amount of labor required to generate ice charts, and the impressive performance of Deep Convolutional Neural Networks (DCNNs) in general-purpose image classification tasks [11], [12], have led to growing interest in the GeoAI community for developing automated sea ice classification algorithms from remotely-sensed images [13]–[21].

In remote sensing imagery analysis, deep learning (DL) significantly enhances both classification and segmentation tasks through its efficient neural architectures. Leveraging multi-layered neural networks, DL models are adept at extracting patterns and features from satellite imagery and have been instrumental in tasks such as land cover/land use classification [22]–[24], object detection [25], and change detection [26]. Models like Convolutional Neural Networks (CNNs) [12] and Recurrent Neural Networks (RNNs) [27], including Long Short-Term Memory (LSTM) networks [28], excel in extracting complex spatial and temporal patterns for classification. For segmentation, architectures such as U-net [29] and DeepLab [30] stand out by delineating precise object boundaries within imagery, enabling detailed environmental monitoring. By automating feature extraction and learning, these DL approaches offer unparalleled accuracy in mapping sea ice in satellite imagery [20], [31], [32].

Supervised deep learning classification algorithms rely on high volumes of high-quality labeled training samples to perform well, and their performance is limited by the quality of the training data [33]. This presents a challenge in many real-world GeoAI and remote sensing applications, including ice classification, where the process of generating accurately labeled training samples from satellite images requires expert knowledge, and is therefore expensive and labor-intensive [12], [13].

Remotely-sensed images (mostly SAR) have been used in traditional machine learning algorithms, such as random forest and Support Vector Machines (SVM), as well as supervised and weakly-supervised deep learning algorithms in multiple applications such as ice-water classification or ice-type classification [8], [18]–[21], [36]–[38], concentration and/or thickness estimation [14], [29], [39]–[41], and sea ice motion prediction [42]. Conventional (non-deep learning) machine learning algorithms, such as random forests and Support Vector Machines (SVM), as well as supervised and weakly-supervised Neural Network (NN)-based algorithms have been widely used for sea ice classification in remote sensing imagery [6]. More recently though, Convolutional Neural Network-based models have been used more frequently due to their ability to capture spatial context and the resulting high performance in general-purpose image classification tasks. With respect to the number of classes for sea ice classification, these studies can be grouped into binary ice-water classification [20], [20], [31], [43], [44], and multi-class ice-type classification [18], [19], [21], [36], [44]–[47].

To generate labels from operational sea ice charts, existing sea ice classification studies approximate pixel- or patch-level labels by encoding SoD as one-hot vectors, either based on the oldest SoD, e.g., in [36], [46], or based on the SoD with the highest partial concentration, e.g., in [47]. Label smoothing, which transforms a one-hot encoded vector of "hard" targets to a vector of soft targets using a uniform distribution [48], is an alternative to one-hot encoding and is shown to improve sea ice classification performance in a multi-class setting [49].

Regardless of the specific approach, one-hot encoding of the labels leads to the loss of valuable information about other SoDs that might be present in a polygon and could potentially lead to incorrectly-labeled samples in the cases where the oldest SoD does not necessarily have the highest partial concentration. Furthermore, to the best of our knowledge, Convolutional Neural Network (CNN)-based sea ice classification studies do not incorporate partial concentration values for classifying sea ice type.

To address the issues related to mixed ice-type polygons, one approach is to only consider pure polygons, i.e., polygons that contain only one ice type or only water [50]. However, even pure polygons may suffer from uncertainties such as containing unintended ice types, which would lead to representativeness error. [51] proposed three different approaches for dealing with uncertainties in the ice chart labels when predicting ice concentrations, namely perturbing the concentration label, augmenting it using SAR data, and using a mean-split loss function. While this is an interesting approach especially for ice concentration, applying it to a multi-class ice-type classification problem is not straightforward. Furthermore, given the limited number of pure polygons, this approach leads to a significant loss of potentially useful training samples.

With regard to the loss function, all CNN-based sea ice classification studies mentioned above use a variation of cross entropy loss (binary or categorical, depending on the number of classes) for optimizing the model. Focal loss, however, has been shown to perform generally better than cross-entropy loss in imbalanced classification settings [52]. While focal loss has been used in sea ice research, it has largely been used in applications other than classification. For instance, [53] used it in a weakly-supervised sea ice segmentation model, [54] used it for sea ice forecasting with probabilistic deep learning based on a U-net architecture [55], and [56] used it for sea ice concentration charting.

When manually classifying sea ice in remotely-sensed imagery, analysts first identify polygons of seemingly homogeneous ice based on the likelihood of the presence of older SoDs (and therefore, thicker ice) within them. Next, they assign up to three different ice types, along with the corresponding partial concentration to each polygon. Partial ice concentration, measured as a percentage, is defined as the relative amount of area (within the polygon) covered by each ice type and determines how much of a polygon is covered by a certain ice type (SoD).

While being more time-efficient, this has some notable drawbacks. Firstly, for practical reasons, such polygons are often large (compared to the spatial resolution of the image) and can contain thousands of pixels [13]. Secondly, an expert might wish, or need to, assign more than one label to a polygon, potentially because of semantic hierarchical categorizations of classes in the domain, or the presence of more than one feature type in the polygon. Finally, the assigned attributes do not determine where exactly in the polygon each ice type resides. The presence of different ice types with different partial concentrations inside a polygon, and the fact that it is not possible to pinpoint the location of individual ice types within the polygon pose great challenges for training conventional supervised sea ice classifiers.

Another common challenge in developing GeoAI models for sea ice classification, and more generally, environmental remote sensing, is that training datasets are often imbalanced meaning that the proportion of training samples in one or more classes is considerably lower compared to other classes. This is a reflection of the uneven distribution of the classes of interest in the real world [57]–[61]. When the training dataset is imbalanced, deep learning classifiers tend to perform well on the more frequent class(es) and underperform on the less frequent–but potentially more important–class(es).

To address the aforementioned challenges, in this paper, we present a novel approach for training sea ice classification algorithms using remotely sensed images that are labeled with polygons containing multiple ice types. We frame this task as a Partial Label Learning (PLL) problem and allow each training sample to have multiple candidate labels (i.e., ice types), only one of which is the true label. We encode the partial concentration of each ice type as a confidence score associated with the corresponding candidate label. This confidence score can also be understood as the probability of that ice type occurring within the polygon. Finally, we integrate these confidence scores within the focal loss function to handle both partial labels and class imbalance[1].

Using this approach, we train a CNN for sea ice classification using dual-polarized Sentinel-1 SAR images in Extra-Wide (EW) swath mode. We optimize and evaluate focal loss hyperparameters for sea ice classification, and compare the performance of our approach to a more conventional single-label learning approach. By improving the accuracy of automated sea ice classification, through the model we propose, we hope to address some of the aforementioned concerns regarding automated sea ice classification approaches. The contributions of this work are as follows:

- A novel framework for training deep learning algorithms on remotely-sensed images with patch-level multi-candidate and imbalanced partial label learning by incorporating partial label learning with focal loss.
- A more efficient deep learning method for sea ice classification based on Sentinel-1 SAR images that improves the performance of the existing methods by incorporating all ice types and partial concentrations from training samples.
- Sensitivity analysis and tuning of hyperparameters of the focal loss function ($\alpha$ and $\gamma$)

The remainder of this paper is structured as follows: Section 2 reviews the related work in the context of PLL and presents our proposed approach and its mathematical formulation as well as the experimental evaluation setup. In Section 3, we present our results and in Section 4 we discuss our findings. Finally, Section 5 presents our conclusions and directions for future work.

## 2. Materials and Methods

### 2.1. Partial Label Learning (PLL)

In this section, we describe our proposed formulation for defining sea ice classification as a partial label learning problem. PLL is a learning paradigm in which each training sample is associated with a set of candidate labels, among which only one is assumed to be the true label [62]. PLL is also known as ambiguous-label learning and superset-label learning [63].

The main goal of PLL is to train a model that identifies the singular true label among a set of candidate labels for an input sample. There are two groups of strategies for doing so: the average-based strategy and the identification-based strategy. In the average-based strategy, all candidate labels are treated equally in the training phase and then the model outputs of all the candidates (which are in the form of probabilities) are averaged for the final prediction [64]. This strategy is simple but has a drawback in that it cannot take the difference between the candidate labels into account and treats them equally. This could be suboptimal in sea ice classification as the candidate labels (different ice types present in a polygon) have different partial concentrations and thus, different likelihoods of being the true label.

In the identification-based strategy, the true label is considered a latent variable, and we assume a parametric model based on which the true label can be identified. For example, the label that achieves the highest probability in the final prediction could be considered the ground-truth label [65], [66]. As such, this strategy covers the gap mentioned for the average-based strategy and is the strategy that we build on in this work.

An important step in leveraging identification-based PLL is to derive the confidence score for each candidate label. Multiple methods have been proposed to estimate and update the confidence scores, for instance, using the information extracted from the feature space to update the label distribution [67], or using norm regularization via self-training [68]. The common characteristic in all of the existing work is that the confidence score is not known a priori, and thus, has to be estimated and updated.

PLL is a branch of Weakly Supervised Learning (WSL). Even though WSL has been used extensively in remote sensing applications, for instance for object detection [69]–[71] and semantic segmentation [72], [73], our work is the first study based on PLL in remote sensing applications to the best of our knowledge.

We formalize sea ice classification as a single-label classification problem and use the partial ice concentration associated with each candidate label as an explicit measure of confidence (or the probability of occurrence of the ice type within the polygon). Partial ice concentrations represent the estimated distribution of different ice types within a polygon (as interpreted by the ice analyst). We use a similar approach as in [74], but further, integrate our approach with focal loss and show that it results in improved performance.

For each training sample, we first generate a binary candidate label vector using the ice types present in the sample (extracted from the containing polygon) and then multiply that vector by a vector of confidence scores element-wise, derived from the partial ice concentration of each type.

We denote $X_{n \times m}$ as the sample (instance) matrix, where n is the number of training samples and m is the number of features in each sample (i.e., the number of channels in an image). The

---

i-th individual training sample is, therefore, $X_i = [x_1, x_2, \cdots, x_m]$.

We then denote the label matrix as $Y_{n \times l} \in \{0,1\}$, such that row $Y_i$ is the label vector corresponding to the $i$-th instance ($X_i$) where $l$ is the number of classes. In a one-hot encoding labeling scheme (widely used in sea ice classification), the ground-truth label for instance $i$ is represented by a 1 in column $j$ and a zero in all other columns (i.e., for instance $i$: $\sum_{j=1}^{l} y_{ij} = 1$). In a partial label learning scheme, however, this is relaxed slightly to allow for multiple **candidate** labels, such that for instance $i$: $1 \leq \sum_{j=1}^{l} y_{ij} \leq l$. Consequently, a partial label for instance $i$ has a 1 in all columns $j$ that are candidate labels and a 0 in all columns $k$ that are non-candidate labels.

The identification-based strategy in partial label learning extends this to incorporate a confidence score associated with each candidate label. That is, for instance $i$ and candidate label $j$, $y_{ij} \in [0,1]$, where a value closer to 1 represents greater confidence in that label being the true label.

We derive the confidence scores from the partial ice concentration associated with each type of ice in operational sea ice charts. Part of the novelty of our approach is that **we assign the partial concentration of an ice type to its probability of being the ground-truth label in a PLL framework** (and then integrate it with the focal loss for sea ice classification). After incorporating partial ice concentration as the confidence score, our label matrix $Y_{pc} \in [0,1]^{n \times l}$ will be of the form:

$$Y_{PC} = Y \otimes C = \begin{bmatrix} y_{11} & y_{12} & \cdots & y_{1l} \\ y_{21} & y_{22} & \cdots & y_{2l} \\ \vdots & \cdots & \ddots & \vdots \\ y_{m1} & y_{m2} & \cdots & y_{ml} \end{bmatrix} \otimes \begin{bmatrix} c_{11} & c_{12} & \cdots & c_{1l} \\ c_{21} & c_{22} & \cdots & c_{2l} \\ \vdots & \cdots & \ddots & \vdots \\ c_{m1} & c_{m2} & \cdots & c_{ml} \end{bmatrix}$$
$$= \begin{bmatrix} y_{pc_1} \\ y_{pc_2} \\ \vdots \\ y_{pc_m} \end{bmatrix} \tag{1}$$

where $Y$ is the binary label matrix, $C$ is the confidence matrix (sea ice concentrations for each instance), and $y_{pc_i}$ is the partial label vector of length $l$ for instance $i$ ($y_{pc_{ij}} \in [0,1]$)[2]. This is similar to the concept of a membership matrix used in [75], but is different in that the values in our vectors represent confidence scores rather than direct membership degrees.

We have demonstrated this in Fig. 1. In the highlighted polygon, which is labeled with two ice types, the oldest ice type or SoD (denoted by SA) is first-year ice which has a partial concentration code 79, representing 70–90% ice (denoted by CA), and the second oldest ice type is young ice (denoted by SB) which has a partial concentration code 24, representing 20–40% ice (denoted by CB). If we sort the ice types by their numeric SA code so that the columns correspond to [NY, SY I, FY I, OI, W], the label vector for this polygon that a one-hot encoding would derive would be $\vec{y}_o = [0, 0, 0, 1, 0, 0]$. In this vector, the value corresponding to the oldest ice type is 1 and the values of all other ice types are zero. The label vector that a binary (conventional) partial labeling would derive is $\vec{y}_P = [0,$

0, 1, 1, 0, 0], where the label is 1 for all ice types present in the polygon and 0 otherwise. With our formulation, the derived label vector would be $\vec{y}_{PC} = [0, 0, 0.3, 0.8, 0, 0]$, where the value of each ice type (class) is the mean of the encoded partial concentration range of that type in the ice charts, as estimated by the expert ice analyst (Fig. 1).

In manually-generated ice charts, the sum of partial ice concentrations for a given polygon may not add up to 100% or even exceed 100% (as seen in this example). To generate partially-encoded label vectors in the latter case, we subtract half of the surplus (over 100) from the partial concentrations of each of the two labels. So $\vec{y}_{PC}$ in the example above will become $\vec{y}_{PC} = [0, 0, 0.25, 0.75, 0, 0]$. It is important to note that we performed this process only during training and not testing (inference). Also, the maximum possible sum of concentrations in our dataset was %110 and only happened when two ice types were present in a polygon.

### 2.2. Focal Loss for Class Imbalance

Similar to many other remote sensing applications of machine learning, sea ice classification is an imbalanced classification task, regardless of the area or the period of study. Class imbalance occurs when one or more classes have much lower proportion of training samples compared to other classes. In the case of sea ice, this can be attributed to the physics of sea ice formation. Younger, thinner ice typically covers relatively small areas and exists for only a short period (a few days) before it either grows into thicker (older) categories or melts away.

To address this issue, we integrate partial labels within the focal loss (FL) function to train our model. Focal loss, first introduced by Lin et al. [52], is a generalization of the Cross-Entropy (CE) loss that is designed to deal with highly imbalanced datasets. CE loss simply calculates the logarithm of the model's estimated probability for each class (often calculated using the Softmax function) and is defined as:

$$CE(p, y) = - \sum_{i=1}^{l} \log(p_i) \cdot y_i \tag{2}$$

where $p_i \in [0, 1]$ is the model's estimated probability for class $i$, $y_i \in \{0,1\}$ is the label for that class, and $l$ is the number of classes. Focal loss adds a scaling (modulating) factor and a focusing parameter to the CE loss. In tandem, these two decrease the loss value for samples from majority class(es), which are often well-classified, and shift the focus to samples from minority class(es), often called hard samples. Focal loss is defined as [52]:

$$FL(p, y) = - \sum_{i=1}^{l} \alpha \ (1 - p_i)^\gamma \ \log(p_i) \cdot y_i \tag{3}$$

where $(1 - p_i)$ is the modulating factor, $\gamma \geq 0$ is the focusing parameter, and $\alpha \in [0, 1]$ is the weighting factor. If $\gamma = 0$ and $\alpha = 1$, focal loss is equivalent to CE loss.

---

[2] We use capital letters for denoting matrices and small letters to denote vectors throughout the paper.

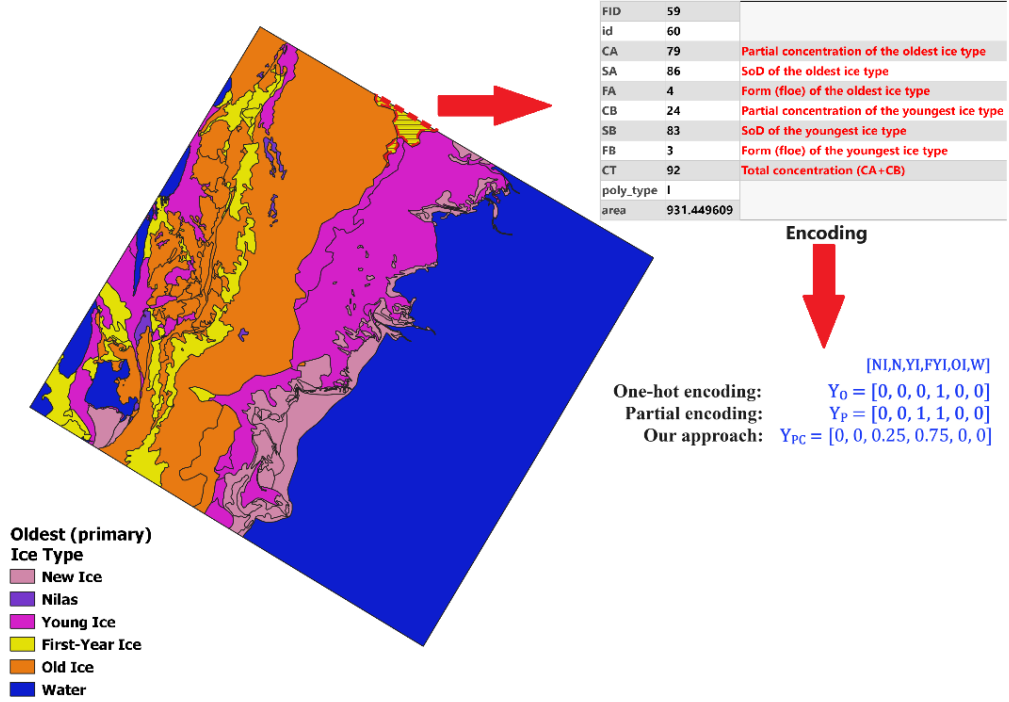

**Fig. 1.** The ice chart generated for the January 16, 2018 image (listed in Supplementary Table I) in the ExtremeEarth dataset with one of its polygons highlighted. Three different encodings of the ice types present in this polygon are demonstrated. Ice type abbreviations: NI: New Ice, N: Nilas, YI: Young Ice, FYI: First-Year Ice, OI: Old Ice, W: Water.

As can be seen in (3), the modulating factor is defined such that it decreases to zero as the probability of the correct class increases, and consequently, reduces the contribution of easy samples to the training loss. The focusing parameter adjusts the rate at which easy samples are down-weighted. Therefore, increasing $\gamma$ would exponentially increase the rate at which such samples are down-weighted and make the model focus on hard samples [75]. The $\alpha$ parameter is used to balance the loss function by adjusting the weight assigned to the rare class(es). It is an optional parameter, but it is shown to improve the performance of the models trained with the focal loss [75]. Both $\gamma$ and $\alpha$ adjust the effect of the rare class, therefore they interact with each other and have to be optimized for a domain application together.

Another approach in dealing with imbalanced datasets in classification problems is to incorporate class weights in the loss function directly. Class weight is calculated as [79]:

$$W_j = \frac{n}{l \times n_j} \qquad (4)$$

where $W_j$ is the weight for class $j$, $n$ is the total number of samples, $l$ is the number of classes, and $n_j$ is the number of samples in class $j$. In this formulation, the weight of each class is inversely proportional to its frequency. Thus, by incorporating (4) in the loss function, minority (less frequent) classes will be weighed higher and therefore the model will be penalized more for misclassifying samples from those classes.

Class weights can be incorporated in both focal loss and cross-entropy loss. The class weights are often multiplied by the individual loss values corresponding to each sample in the training batch. By doing so, the loss function gives more importance to the minority class samples, effectively addressing the imbalance issue and helping the model learn from the rare classes. Therefore, in addition to evaluating different label encoding methods, we also incorporate class weights into both loss functions and evaluate the performance of the models with and without those weights for sea ice classification.

### 2.3. Model Architecture

In our preliminary research, we compared the performance of four CNN-based image classification architectures for the task of sea ice classification and identified the pre-trained version of ResNet-50 as the best-performing architecture [80]. The superiority of ResNet can likely be attributed to its use of skip connections and residual blocks. These features enable it to deliver enhanced performance with a lower number of trainable parameters (approximately 25.6 million in ResNet-50) compared to models like VGG-16, which has approximately 138 million trainable parameters. In our context, where we deal a relatively small dataset, having fewer trainable parameters is advantageous. Therefore, in this work, we use the same architecture as the backbone of the model and freeze the pretrained weights of the convolutional layers (pre-trained on the ImageNet dataset [81]). Employing a carefully-selected pretrained neural network can partly alleviate the challenge of having a small number of training samples, which is the case in our experiments. To fine-tune the model for sea ice classification, we add two fully connected layers of 64 neurons, each followed by a batch-normalization layer [82], ReLU activation function, and dropout [83]. Dropout regularizes the model and reduces its generalization error, and thus, helps prevent overfitting the model to the training samples. We experimented with different dropout rates and found 0.25 to yield the best performance (more detail in Section 2.4). We finally added an output layer with 6 neurons (equal to the number of classes) and Softmax activation. Fig. 2 presents the architecture of our model.

Many pre-trained image classification models, including the ResNet, take three-channel RGB images (such as those provided in the ImageNet dataset) as input. Sentinel-1 SAR records backscatter in HH and HV polarization, and therefore, the resulting images have two channels by default (one per polarization). To create a third channel, others have used different linear combinations of HH and HV channels. For instance, [33] used HV, HH-HV, and HH/HV as the three channels to create pseudo-RGB images. However, this approach may not introduce new information, given the existing presence of the two channels in the input. In this work, we use local incidence angle measurements as the third channel, since incidence angle is shown to affect backscatter intensity differently depending on ice type [44], and thus, can provide additional information to the classifier. To facilitate reproducibility and reuse in the remote sensing research community, an open-source implementation of our approach is available on GitHub at https://github.com/geohai/PLL-sea-ice-classification.

### 2.4. Experimental Evaluation

In this section, we introduce the dataset and then describe the experiments, the parameters used in each, and the results.

#### 2.4.1. Dataset

We use the "ExtremeEarth Polar Use Case Training Dataset Version 2.0.0" for our experiments. This dataset covers the Danmarkshavn region east of Greenland and contains sea ice charts for 12 Sentinel-1 SAR images, each acquired approximately one month apart in 2018 [76]. It was generated as part of the multi-institutional ExtremeEarth project [4] and is designed to serve as a training or validation dataset for automated satellite image processing algorithms. As a result, it provides high-resolution ice charts that would not be publicly available otherwise. To the best of our knowledge, this dataset has not been used for sea ice classification research before.

The ice analysts from the Norwegian MET have interpreted the imagery in the dataset to draw ice charts. The generated ice charts include 6 different ice types: new ice, nilas, young ice, first-year ice, old ice, and ice-free (or water). New ice describes ice that has been recently formed and has a thickness below 10 cm. Nilas is a thin sheet of smooth ice that has a similar thickness to new ice but visually looks darker (especially when thin).

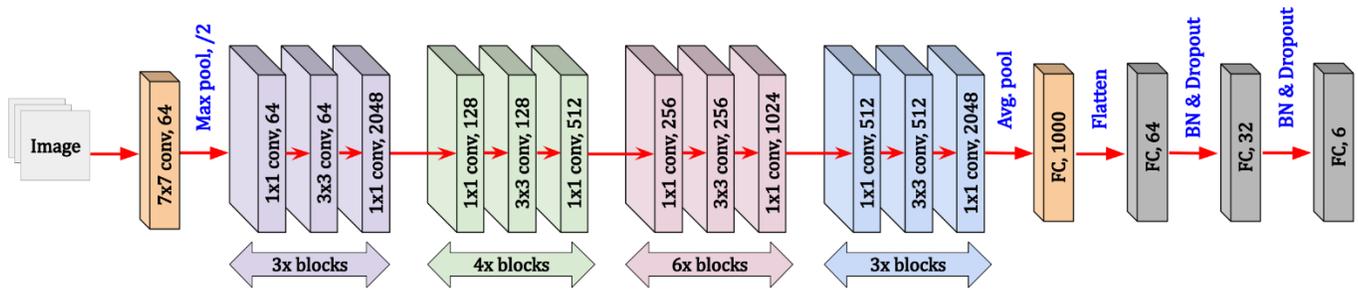

**Fig. 2**. Model architecture based on ResNet-50 model with added fully connected layers. The ResNet-50 architecture includes an initial 7x7 convolutional layer, max pooling, and multiple residual blocks with 1x1 and 3x3 convolutions. Batch normalization (BN) is applied after each convolution, and ReLU activation functions introduce non-linearity. Skip connections bypass layers to improve training, followed by global average pooling and a fully connected (FC) layer. In a pretrained model, all the layers up to "FC, 1000" use the weights pretrained on the ImageNet dataset. We added 3 fully connected layers with batch normalization (BN) and dropout for fine tuning. The final layer outputs 6 classes.

Young ice refers to ice that has a thickness between 10 and 30 cm. First-year ice is ice that has formed since the onset of freeze-up and has not survived a summer melt season. Its thickness is greater than 30 cm and typically grows up to 2 m (depending on air temperatures and time since formation). Sea ice that has survived at least one melt season is called old ice (also multi-year or perennial ice) and usually has a thickness above 2 m. The corresponding code used for representing these ice types in the ice charts is 81, 82, 83, 86, and 95 respectively. Monitoring the ice type, and therefore its thickness is crucial for marine navigation as thicker ice types may present hazards for vessels depending on their ice-breaking grade.

During the melt season, melt ponds form on the surface of the ice. Additionally, the snow on the surface of the ice melts and creates a wet (moist) layer. The melt ponds and the wet snow alter the radar backscatter and make automated classification prone to error, especially when using C-band SAR, as is the case with Sentinel-1 SAR acquisition. Therefore, we focus on the freeze-up season in the Arctic for this study and only use the 6 images acquired in January, February, March, October, November, and December of 2018. The acquisition times and the file names of these images are listed in Supplementary Table I and their footprint is presented in Fig. 3.

The Sentinel-1 images in this dataset were all acquired in the Extra-Wide (EW) swath mode with HH and HV co- and cross-polarized channels. To process the raw images, we first apply radiometric and orbital corrections on each image, replicating the processes that are described in the dataset documentation [76]. To minimize the effect of thermal noise on the images, we also apply the thermal noise removal algorithm provided in the Sentinel Application Platform (SNAP)[3]. This algorithm, however, does not sufficiently remove the noise in the HV polarization (see Fig. 4b).

The residual noise can be misclassified as ice artifacts by automated algorithms [36], thus, we manually mask patches with high residual noise. Fig. 4 presents the HV cross-polarized channel of the image acquired in January in addition to the denoised version of this channel, as well as the resulting image after masking noisy patches.

The spatial footprint of these images is approximately 400km × 400km. With a pixel spacing of 40m×40m, each image contains approximately 10,000×10,000 pixels. To prepare samples and labels for a CNN model, we divide each image into 50 × 50-pixel patches, each covering a 2km × 2km area on the ground. We then align each image with its corresponding ice chart (by projecting it into the same projection system as the ice chart), overlay the patches with the ice polygons, and store the information about the ice types present in the patch once as one-hot encoded labels, and once as partial label vectors with confidence (to be used in the experiments outlined below). To perform a robust evaluation of the models, we only consider samples where the concentration of the oldest ice type is above 50%, or in other words, the samples where the oldest ice type is also the most dominant. In addition, to avoid the potential inaccuracies in labeling areas close to ice polygon borders [36], we excluded the samples where the distance from the center of the patch to any polygon border was less than 2 km.

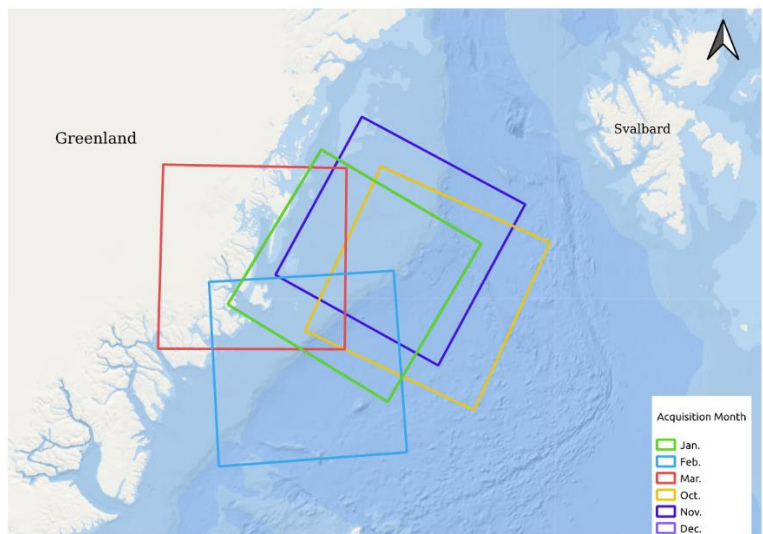

**Fig. 3**. The footprint of the six Sentinel-1 images used in our experiments. Note that the January and December images have identical footprints that are overlapped, thus there appears to be 5 footprints on the image. Map scale: 1:5,000,000.

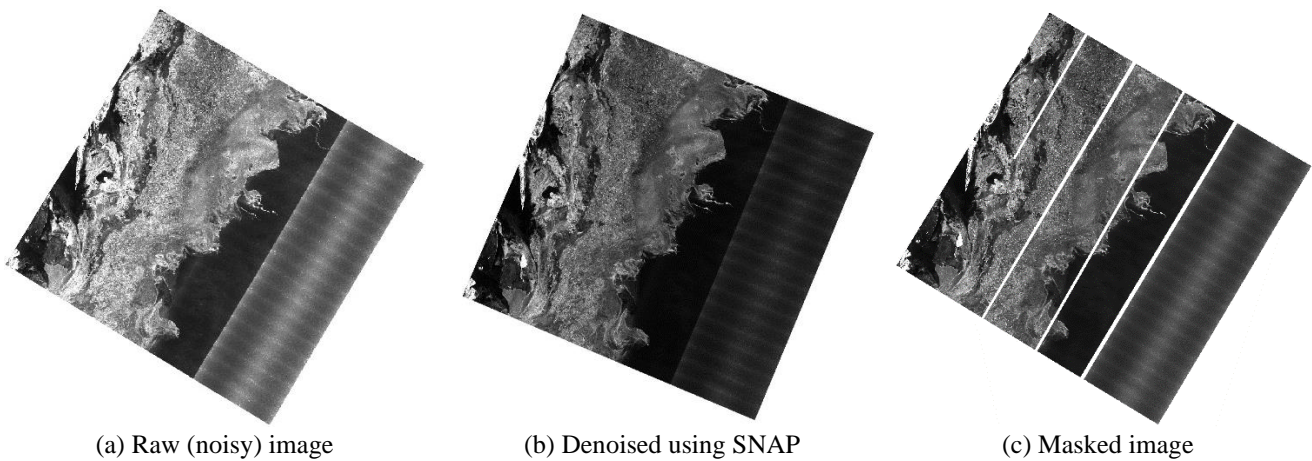

(a) Raw (noisy) image       (b) Denoised using SNAP       (c) Masked image

**Fig. 4**. The HV cross-polarized channel of the January image (file name and acquisition time listed in Supplementary Table I). (a) shows the raw backscatter values and (b) shows the denoised version (using SNAP). Note the amount of residual noise still present

---

[3] https://earth.esa.int/eogateway/tools/snap

in (b), especially along sub-swath edges. To remove the residual noise, we manually excluded pixels along the sub-swath boundaries in all images. (c) shows the image after manual masking.

### 2.4.2. Experiment Design

We hypothesize that our proposed approach in encoding ice charts as confidence-aware partial labels could outperform traditional one-hot encoding alternatives in training sea ice classification. This is because our approach capitalizes on all the available information in samples when training the network. Additionally, we posit that integrating confidence-aware partial labels with the focal loss will effectively address the class imbalance issue.

To investigate these hypotheses, we designed a set of experiments to compare the performance of two types of training label encoding: partial and one-hot encoding. Concurrently, these experiments allow us to evaluate two strategies for mitigating class imbalance: utilizing CCE loss with class weights or applying focal loss (FL).

The experiments are divided into two groups based on the loss function: the first group utilized the CCE loss, while the second deployed the FL. Within each group, we independently used partial encoded labels and one-hot-encoded labels to train separate models, enabling us to scrutinize the impact of label encoding on model performance. In order to identify the most effective method for handling class imbalance (outlined in Section 2.2), we conducted each set of experiments in group 1 with and without class weights. Table I presents a summary of these experiments.

Furthermore, we analyze the sensitivity of the focal loss to the $\alpha$ and $\gamma$ hyperparameters for sea ice classification. As such, we adopt $\{0.1, 0.25, 0.5, 0.75, 0.9\}$ as potential $\alpha$ values, and $\{1, 2, 5\}$ as potential $\gamma$ values. For each set of experiments in group 2 and for each unique combination of these values, we train a separate model.

We performed a total of 34 experiments (Table I), each corresponding to a CNN model, with different configurations to reflect the setting of the experiment. We use the Adam method [77] and a learning rate of $10^{-3}$ for optimizing the models and train them for 200 epochs with a batch size of 512. We developed the models using the PyTorch library [78] and trained them on a system with dual Nvidia RTX A5000 GPUs, which allowed for a training time of 20-25 seconds per epoch.

TABLE I
A SUMMARY OF EXPERIMENTS

| Group | Loss Function | Training Labels | Variable | Values | # of models |
|---|---|---|---|---|---|
| 1 | Categorical Cross Entropy | Partial | Class weights | With and w/o class weights | 2 |
| | | One-hot-encoded | Class weights | With and w/o class weights | 2 |
| 2 | Focal | Partial | Alpha | {0.1, 0.25, 0.5, 0.75, 0.9} | 15 |
| | | | Gamma | {1, 2, 5} | |
| | | One-hot-encoded | Alpha | {0.1, 0.25, 0.5, 0.75, 0.9} | 15 |
| | | | Gamma | {1, 2, 5} | |

### 2.4.3. Evaluation

In line with the typical approach in deep learning evaluation, we randomly divided our dataset of 127K samples into training, validation, and testing subsets. The ratios for these subsets were set at 81%, 9%, and 10% respectively. To evaluate the performance of our models, we utilized the unseen test set and computed the following metrics:

$$\text{Accuracy} = \frac{TP+TN}{TP+FP+TN+FN} \quad (5)$$

$$\text{Weighted Average Precision} = \sum_{i=1}^{l} \left( \frac{n_i}{n} \times \frac{TP_i}{TP_i+FP_i} \right) \quad (6)$$

$$\text{Weighted Average Recall} = \sum_{i=1}^{l} \left( \frac{n_i}{n} \times \frac{TP_i}{TP_i+FN_i} \right) \quad (7)$$

$$\text{Weighted Average F-1 score} = \sum_{i=1}^{l} \left( \frac{n_i}{n} \times \frac{2TP_i}{2TP_i+FP_i+FN_i} \right) \quad (8)$$

$$\text{Per-class F-1 score} = 2 \times \frac{Precision_i \times Recall_i}{Precision_i + Recall_i} = \frac{2TP_i}{2TP_i+FP_i+FN_i} \quad (9)$$

where $n$ is the total number of samples, $l$ is the number of classes (6 in our experiments), and $n_i$, $TP_i$, $TN_i$, $FP_i$, and $FN_i$ are the number of samples, True Positive predictions, True Negative predictions, False Positive predictions, and False Negative predictions, respectively for a given class $i$.

As seen in (6)-(8), weighted average precision, recall, and F-1 score are weighted by the number of samples (or support) in each class, and as such, are suitable for our imbalanced classification task. F-1 score is the harmonic mean of precision and recall which takes the number of prediction errors as well as the type of such errors into account. Therefore, in an imbalanced classification task, F-1 provides a more robust measure of model performance compared to accuracy. Thus, we use the weighted average F-1 score as the primary metric to compare our models.

To reduce the stochasticity associated with neural network-based models, we repeated each experiment twice and reported the average values for each metric. Even though we did not notice a large variation, we acknowledge that a higher number of repetitions could result in more robust estimates. In addition to these evaluation metrics, we compare the convergence speed of the models.

## 3. RESULTS

### 3.1. Experiment Group One: Categorical Cross-Entropy Loss

This experiment group consists of models trained with CCE loss, one-hot or partial labels, and weighted or unweighted samples. As seen in Table II, leveraging partial labels, instead of one-hot encoded labels, leads to higher weighted average F-1 scores, regardless of whether class weights are incorporated in the loss function. It also leads to better or equal performance across other aggregated metrics. Moreover, the scores are less sensitive to the use of class weights when partial labels are used. However, the results are not conclusive in terms of per-class F-1 scores, as using partial labels leads to better F-1 scores in half of the classes (three of the six in both experiments).

On the other hand, regardless of the method used for encoding the labels, integrating class weights into the CCE loss yields lower weighted average F-1 scores. Furthermore, it results in lower per-class F-1 scores for all classes except young ice, and lower performance across all other aggregated metrics.

Additionally, the label encoding method makes little difference in convergence speed when class weights are not used in minimizing CCE Loss (Fig. 5). This is corroborated by the performance values in the second and fourth rows of Table II. Therefore, even though using confidence-aware partial labels in tandem with CCE loss leads to improvements in classification performance, these improvements are marginal.

In our initial experiments, the models in this group did not seem to converge after 200 epochs. Therefore, to generate Fig. 5 we trained these models for 300 epochs so that they can achieve a stable state. It is important to note though, that the results in Table II are generated after 200 epochs of training in order to keep the comparison between models of different groups consistent.

TABLE II

PERFORMANCE RESULTS OF THE MODELS TRAINED WITH CCE LOSS. ALL METRICS, EXCEPT FOR TRAINING ACCURACY, ARE MEASURED ON THE UNSEEN TEST SET.

| Encoding | Class Weights | Training Accuracy | Test Accuracy | Weighted Average | | | Per-class F-1 Score [NI, N, YI, FYI, OI, W] |
|---|---|---|---|---|---|---|---|
| | | | | F-1 | Precision | Recall | |
| One hot | Yes | 63.77 | 64 | 59 | 58 | 64 | [35, 7, 51, 28, 84, 89] |
| | No | 88.85 | 87 | 90 | 93 | 87 | [89, 13, 49, 63, 84, 91] |
| Partial | Yes | 76.91 | 68 | 76 | 91 | 68 | [83, 4, 44, 68, 86, 83] |
| | No | 89.68 | 89 | 91 | 93 | 89 | [87, 15, 44, 66, 83, 96] |

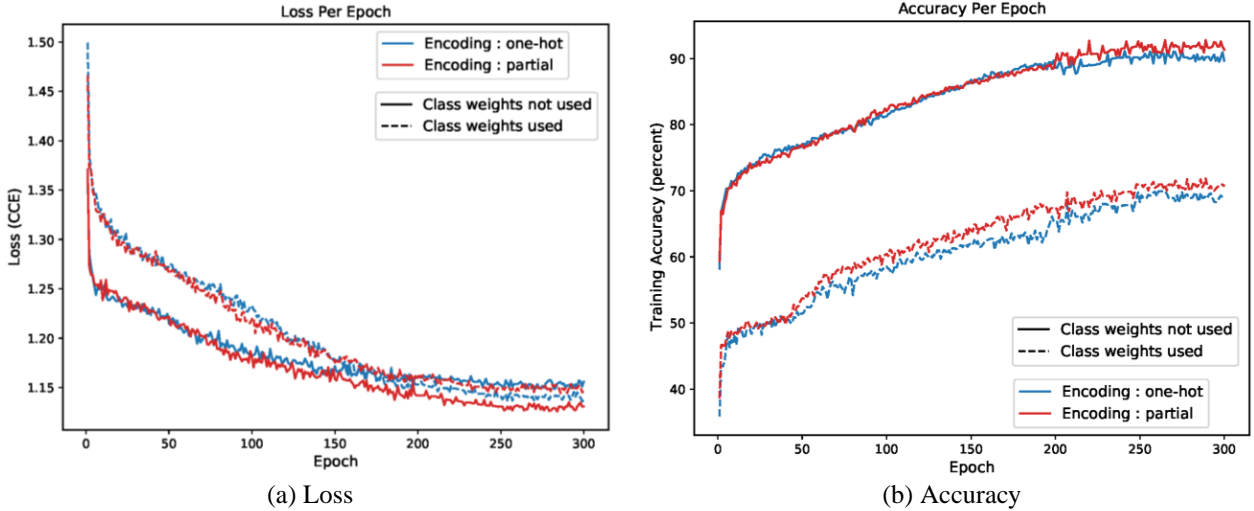

(a) Loss                                          (b) Accuracy

**Fig. 5.** Training loss and accuracy plots for the first group of experiments (trained with CCE loss).

### 3.2. Experiment Group Two: Focal Loss

Table III presents the performance of the 15 models trained with confidence-aware partial labels integrated with focal loss, with various values of hyperparameters. The model with $\alpha = 0.25$ and $\gamma = 1$ achieved the highest weighted average F-1 score among all the 34 experiments we carried out across all groups, confirming our hypothesis that integrating partial labels with focal loss achieves the best performance for training sea ice classification using ice charts as labels. The model with $\alpha = 0.25$ and $\gamma = 1$ also achieved the highest per-class F-1 score in 3 out of the 6 classes (young ice, first-year ice, and water), as well as the highest test accuracy and weighted average recall and precision (with precision tied with 5 other models).

Furthermore, this model achieved a better per-class recall in 4 of the 6 classes compared to the model trained with CCE loss and partial labels, and thus, provides a more balanced classification as all classes have a recall of at least 60% (Fig. 7).

Compared to the model trained with CCE loss and 'Class Weights=No' (the second row in Table II), our best-performing model achieves improvements in the following metrics: Test accuracy: 5%, weighted average F-1: 3%, weighted average recall: 5%, per-class F-1 score for the following classes: nilas: 6%, young ice: 4%, first-year ice: 5%, and water 7%.

However, not all models in this group achieve better weighted average F-1 scores compared to the model trained with CCE loss and confidence-aware partial labels (the last row in Table II). This points to the importance of tuning the hyperparameters of the focal loss for the specific task, in this case, sea ice classification. When its hyperparameters are tuned, using focal loss yields higher F-1 scores (weighted average and per class) compared to the CEE loss.

Due to space limitations, Fig. 6 only presents the training loss and accuracy for four of the models from Table III with $\alpha \in \{0.25, 0.5\}$ and $\gamma \in \{1, 5\}$. Comparing this figure with Fig. 5, we can observe that the models trained with focal loss converge to a stable condition faster than the models trained with CCE loss.

Table III (and Fig. 6b) shows that when $\gamma = 1$, different $\alpha$ values yield similar training accuracies. Despite a similar performance in training, the test metrics vary as the value of $\alpha$ changes. This points to the better generalizability of some models (for instance $\alpha = 0.25$) compared to others (for instance $\alpha = 0.75$) and potentially less overfitting. This can be explained by reviewing (3), where the value of the focal loss when $\gamma = 1$ is determined by $\alpha$ and $(1-p_i)$, or the modulating factor, and the interaction between these two terms.

Using the same $\alpha$ and $\gamma$ values as those used in Table III, we trained additional models with one-hot encoded labels to examine the effects of label encoding when integrated with focal loss (Table IV). Comparing each model with its counterpart in Table III, we can observe that using confidence-aware partial labels leads to improvement in weighted average F-1 score and test accuracy in 9 out of the 15 experiments. More specifically, we can see that when $\alpha \in \{0.25, 0.5\}$, the models perform generally better. Also, within the same range of values for $\alpha$, models trained with confidence-aware partial labels outperform their one-hot encoded counterparts in 5 out of 6 experiments. This is further evidence of the importance of optimizing the hyper-parameter of the focal loss function with respect to the label encoding being used.

Fig. 7 presents the confusion matrices for both our highest-performing model and the traditional model used for sea ice classification, which uses one-hot encoded labels but does not weigh the loss with class weights. Although the conventional model is more adept at classifying new ice and young ice than our model, it struggles to achieve over 50% recall when identifying nilas and first-year ice. This result suggests that the traditional CCE approach may lack consistency when training sea ice classification models.

TABLE III
PERFORMANCE RESULTS OF THE MODELS TRAINED WITH CONFIDENCE-AWARE PARTIAL LABELS INTEGRATED WITH FOCAL LOSS. ALL METRICS, EXCEPT FOR TRAINING ACCURACY, ARE MEASURED ON THE UNSEEN TEST SET. THE MODEL WITH THE BEST WEIGHTED AVERAGE F-1 SCORE AMONG ALL GROUPS IS BOLDFACED.

| Alpha | Gamma | Training Accuracy | Test Accuracy | Weighted Average | | | Per-class F-1 Score [NI, N, YI, FYI, OI, W] |
|---|---|---|---|---|---|---|---|
| | | | | F-1 | Precision | Recall | |
| 0.1 | 1 | 98.79 | 72 | 77 | 88 | 72 | [48, 8, 26, 52, 79, 84] |
| | 2 | 98.44 | 80 | 83 | 89 | 80 | [73, 11, 29, 50, 80, 90] |
| | 5 | 91.05 | 68 | 75 | 89 | 68 | [50, 3, 38, 39, 80, 82] |
| 0.25 | 1 | **98.91** | **92** | **93** | **93** | **92** | **[89, 19, 53, 68, 82, 98]** |
| | 2 | 98.04 | 85 | 88 | 92 | 85 | [84, 10, 33, 65, 82, 93] |
| | 5 | 92.47 | 89 | 91 | 93 | 89 | [87, 18, 47, 61, 81, 96] |
| 0.5 | 1 | 98.98 | 89 | 90 | 92 | 89 | [89, 24, 49, 58, 76, 96] |
| | 2 | 98.58 | 83 | 87 | 91 | 83 | [81, 13, 33, 61, 83, 92] |
| | 5 | 94.00 | 84 | 87 | 92 | 84 | [82, 7, 38, 61, 83, 93] |
| 0.75 | 1 | 99.29 | 85 | 88 | 92 | 85 | [86, 14, 35, 61, 81, 93] |
| | 2 | 98.81 | 86 | 88 | 92 | 86 | [87, 12, 36, 62, 82, 94] |
| | 5 | 94.80 | 81 | 84 | 90 | 81 | [72, 13, 36, 57, 83, 90] |
| 0.9 | 1 | 99.44 | 89 | 91 | 93 | 89 | [87, 12, 45, 64, 81, 96] |
| | 2 | 99.08 | 89 | 90 | 92 | 89 | [88, 12, 39, 61, 82, 96] |
| | 5 | 93.87 | 73 | 79 | 89 | 73 | [60, 6, 36, 55, 82, 84] |

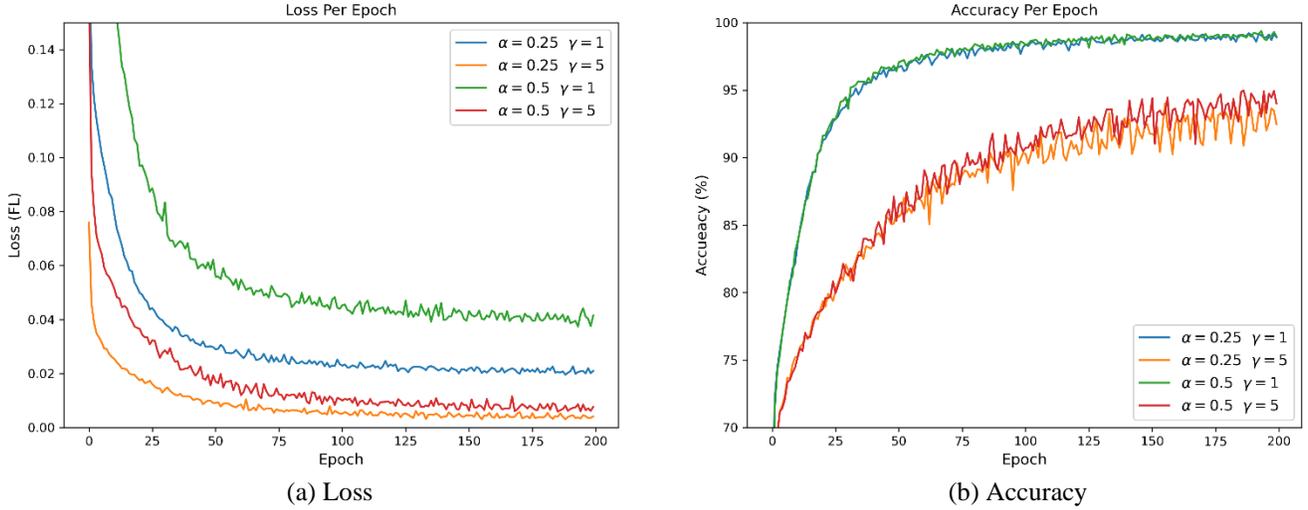

(a) Loss

(b) Accuracy

**Fig. 6.** Training loss and accuracy plots for four of the models trained with confidence-aware partial labels and focal loss ($\alpha \in \{0.25, 0.5\}$ and $\gamma \in \{1, 5\}$).

It is important to note that while the best model in Table IV ($\alpha = 0.25$, $\gamma = 5$) achieves a weighted F-1 score within 2% of the best model in Table III, the performance of the models trained with one-hot encoded labels and focal loss (Table IV) as a whole are less consistent compared to those trained with confidence-aware partial labels. This suggests that leveraging confidence-aware partial label encoding is advantageous in the context of robust sea ice classification.

TABLE IV

PERFORMANCE RESULTS OF THE MODELS TRAINED WITH ONE-HOT ENCODED LABELS INTEGRATED WITH FOCAL LOSS. ALL METRICS, EXCEPT FOR TRAINING ACCURACY, ARE MEASURED ON THE UNSEEN TEST SET.

| Alpha | Gamma | Training Accuracy | Test Accuracy | Weighted Average | | | Per-class F-1 Score [NI, N, YI, FYI, OI, W] |
|---|---|---|---|---|---|---|---|
| | | | | F-1 | Precision | Recall | |
| 0.1 | 1 | 94.11 | 85 | 87 | 92 | 85 | [83, 13, 33, 59, 82, 94] |
| | 2 | 97.37 | 82 | 83 | 85 | 82 | [83, 12, 36, 60, 79, 91] |
| | 5 | 93.22 | 84 | 86 | 80 | 84 | [64, 18, 47, 52, 80, 89] |
| 0.25 | 1 | 99.07 | 84 | 86 | 90 | 84 | [78, 10, 37, 56, 81, 93] |
| | 2 | 98.88 | 84 | 87 | 92 | 84 | [86, 8, 46, 60, 81, 94] |
| | 5 | 94.72 | 89 | 91 | 93 | 89 | [90, 13, 50, 65, 82, 96] |
| 0.5 | 1 | 99.41 | 86 | 88 | 91 | 86 | [88, 12, 36, 59, 81, 94] |
| | 2 | 98.94 | 87 | 88 | 91 | 87 | [64, 24, 48, 62, 83, 95] |
| | 5 | 93.29 | 80 | 84 | 90 | 80 | [78, 12, 33, 52, 80, 90] |
| 0.75 | 1 | 99.51 | 67 | 72 | 87 | 67 | [39, 6, 36, 46, 82, 79] |
| | 2 | 99.47 | 88 | 89 | 91 | 88 | [85, 17, 39, 64, 79, 95] |
| | 5 | 97.52 | 82 | 86 | 92 | 82 | [83, 7, 42, 59, 83, 91] |
| 0.9 | 1 | 99.34 | 88 | 89 | 92 | 88 | [87, 15, 35, 58, 79, 96] |
| | 2 | 99.49 | 81 | 84 | 90 | 81 | [77, 17, 30, 53, 82, 90] |
| | 5 | 98.34 | 84 | 87 | 91 | 84 | [87, 20, 33, 53, 79, 93] |

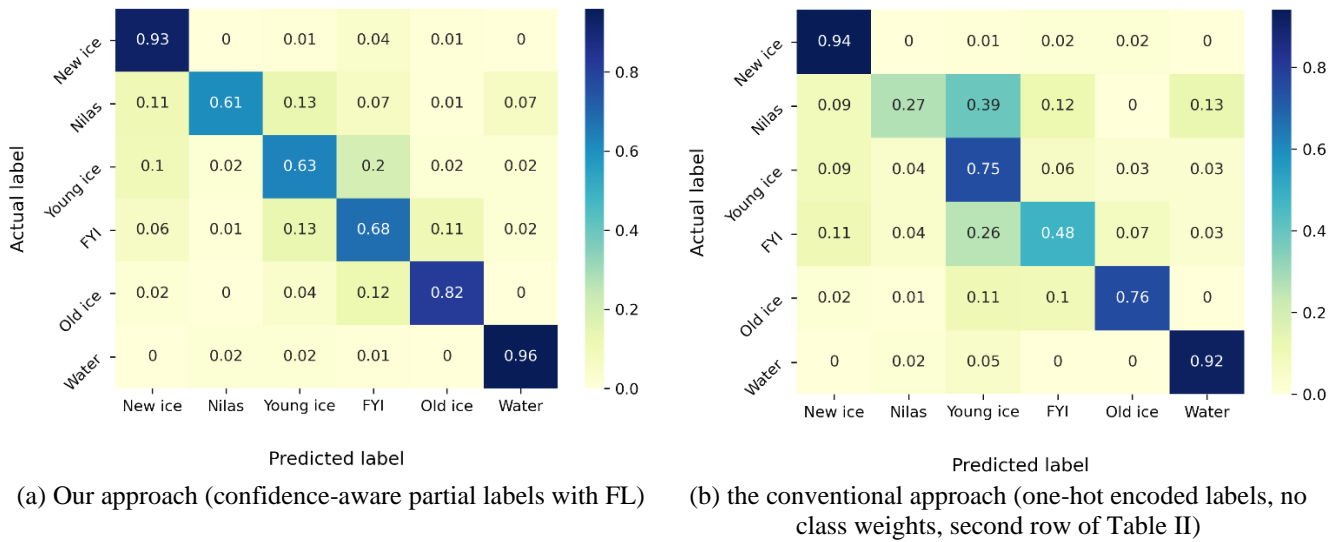

(a) Our approach (confidence-aware partial labels with FL)  (b) the conventional approach (one-hot encoded labels, no class weights, second row of Table II)

**Fig. 7.** Confusion Matrices. a) the model trained with confidence-aware partial labels and focal loss ($\alpha$=0.25 , $\gamma$=1). b) the conventional approach in which we used one-hot encoded labels but did not weigh the loss with class weights.

### 3.3. Focal Loss Sensitivity Analysis

As the results presented in Section 3.2 show, the $\alpha$ and $\gamma$ hyperparameters of focal loss should be tuned to utilize its full potential for sea ice classification using Sentinel-1 images. $\alpha$ adjusts the weight assigned to the rare class(es) and $\gamma$ determines how much the loss should be down-weighted for well-classified examples, therefore, their optimal values depend on the training dataset and, in particular, on the degree of imbalance and difficulty in classification among classes. For instance, Supplementary Fig. 3 shows that nilas and young ice classes have a higher sensitivity to $\gamma$ values. This could primarily be explained by the lower frequency of samples from these classes in our dataset, which is also inherent in sea ice type classification in general. The sensitivity to $\gamma$ values, however, is not necessarily a problem, and rather, shows the importance of tuning the focal loss for hyperparameters for sea ice classification to achieve optimal performance. The $\alpha$ and $\gamma$ hyperparameters tune how much attention the model should pay to the more difficult and less frequent classes (compared to the easier, more frequent classes). The results of our sensitivity analysis, presented in Supplementary Fig. 4, show that $\gamma$ = 1 or 2 work similarly well across all $\alpha$ values, but $\gamma$ = 5 consistently underperforms in comparison. Furthermore, the optimal range of values for $\alpha$ seems to be $\in$ [0.25, 0.75], as values outside of this range result in lower stability and lower performance, especially on the test set. Table III further proves this point by providing the performance of each combination of values on the test set. Our findings show that the combination of $\alpha$ = 0.25 and $\gamma$ = 1 yields the optimal results in sea ice classification.

### 3.4. Visual Inspection of Out-of-Distribution Samples

Considering the number of scenes available in the ExtremeEarth dataset, leaving an entire scene out (from the freeze-up season) for testing would lead to an approximate 20% reduction in the already-limited number of training samples. It would also mean that the model would not have seen any samples from that specific freeze-up month during training.

Therefore, we decided to split the entire sample set into train, evaluation, and unseen test sample subsets as mentioned before. The results provided in Section 3.2 demonstrate that with a test accuracy of 92% and a weighted F-1 score of 93%, our best-performing model can generalize well to unseen test samples. This can be interpreted as the generalization power of the model when extrapolating spatially, as the test samples were randomly chosen from one of the 6 scenes.

To further evaluate the generalizability of our model, we use it to classify an entirely unseen image from the melt season and compare its performance against a benchmark model. The benchmark model is the model presented in the second row of Table II, which has an identical architecture to our model, but is trained with one-hot encoded labels and CCE loss which is the conventional approach in training CNNs on ice charts.

It is important to note that since our model is only trained for the freeze-up season, samples taken from the melt season can be considered out-of-distribution samples. This is exacerbated by the differences in sea ice dynamics and SAR backscatter across different months and seasons, especially between the freeze-up and melt seasons. Therefore, we do not expect the models to achieve high accuracies when generating predictions for images acquired during the melt season. Yet, this experiment could provide visual insight into the generalizability of our proposed approach compared to the conventional approach with limited training data in the benchmark dataset.

To perform this experiment, we chose the September scene as the month immediately preceding the training period. Based on the information provided in the ExtremeEarth dataset user instructions [76], the surface air temperature had ranged between -5° C and 5° C prior to the acquisition of the September image, which in many areas of the image is above the typical -1.8° C freezing temperature that sea ice formation begins. Therefore, we can assume that it is acquired towards the end of the melt season.

Fig. 8 provides a visual comparison between the two models when classifying the September scene. As can be seen in this figure, neither model performs remarkably well in identifying the primary ice types (Fig. 8 c,g). So, we also compared the model predictions with both the primary and secondary ice types (SA and SB) in the label ice charts. In this case, we assume model predictions to be correct if they are equal to either SA or SB classes provided in the ice charts. When model predictions are compared to the primary ice type label (predicted label=SA), the benchmark model achieves an accuracy of 36% whereas our model achieves an accuracy of 57%. Alternatively, when model predictions are compared to either the primary or the secondary ice type (predicted label=SA or SB), the accuracies of the benchmark model and our proposed approach are 42% and 62% respectively.

This shows that our approach in training a deep learning model outperforms the conventional benchmark by achieving 19% and 20% improvements respectively. It is important to note that these metrics are generated using all the samples from the image (not just samples where CA>50%).

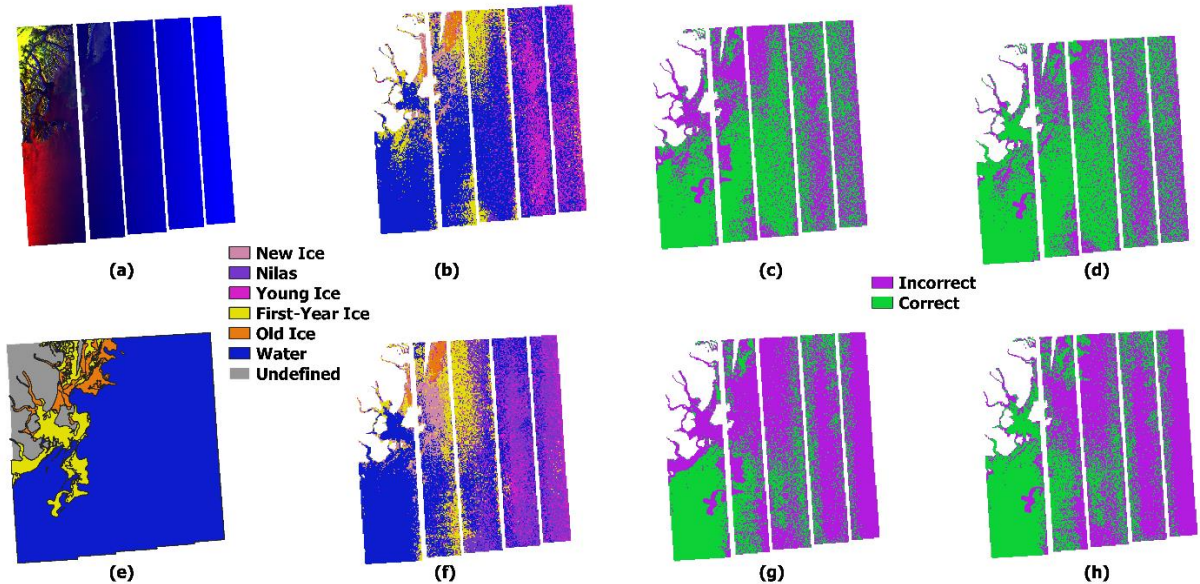

**Fig. 8.** Model performances in predicting and entirely unseen Sentinel-1 image from ExtremeEarth dataset acquired in Sept. 2018. (a) Sentinel-1 SAR image (R: HH channel, G: HV channel, B: local incidence angle. Subswath edges are manually masked out. (b) sea ice types generated by our best-performing model (c) prediction errors in b against label SA values (d) prediction errors in b against label SA and SB values. (e) ice chart label provided in the ExtremeEarth dataset color-coded using SA. (f) sea ice types generated by the benchmark model (trained with one-hot encoded labels as CCE loss) (g) prediction errors in f against label SA values (h) prediction errors in f against label SA and SB values. The results are highly affected by banding noise in the images. The number of incorrectly predicted patches is reduced when using partial concentrations integrated with focal loss and further reduced if either SA or SB are taken as correct labels.

### 3.5. Comparative Analysis

This section presents a comparative analysis of our results with similar sea ice classification studies. Our selection criteria for including these studies are as follows: 1) the study must classify multiple sea ice types (as opposed to binary ice-water classification or ice concentration classification), 2) it should exclusively use Sentinel-1 images as input, and 3) it should employ supervised classification. [36] considered four classes in their study (ice free, young ice, first-year ice, and old ice) and evaluated their model on two datasets. They achieve accuracies of 90.5% on a dataset of 2018 images and 91.6% on a dataset of 2020 images. However, to the best of our knowledge, these accuracies are measured on a validation set as opposed to a test set. Our model achieved a test accuracy of 92% which is higher than both, but underperformed the model developed by [36] in per-class accuracy of young ice (0.77 vs 0.63), first-year ice (0.85 vs 0.68), old ice (0.98 vs 0.82), and water (0.98 vs 0.96). This could be due to the higher number of classes in our model which makes the classification task more challenging, as well as the finer labels present in our dataset. We also replicated the architecture used in [36] following the parts of code available on GitHub and using the same hyperparameters as those listed in the paper and tested it on our dataset. This model achieved a test accuracy of 70.00%, an average F-1 score of 72.3%, and per-class F-1 scores of 58.4% for new ice, 9.8% for Nilas, 42.5% for young ice, 38.6% for FYI, 73.4% for old ice, and 90.7% for water. These are lower compared to our best model as well as the model trained with partial labels in group one (with CCE loss). It should be noted that the model proposed by [36] was trained only using HH and HV channels and has approximately 72K trainable parameters (compared to 25M parameters in our model).

Using the same 2018 dataset mentioned above, [44] trained random forest classifiers using three different Feature Configurations (FC), all of which included texture features derived from Sentinel-1 images. In five-class classification (open water, new ice, young ice, first-year ice, and old ice), their model achieved an accuracy of 60% in winter, which is considerably lower than our model. Their highest per-class accuracy (depending on the FC) in four of the five classes was lower compared to our model (0.45 vs 0.93 for new ice, 0.626 vs 0.63 for young ice, 0.659 vs 0.68 for FYI, and 0.924 vs 0.96 for water), but their accuracy for old ice was higher than ours (0.906 vs 0.82).

[79] developed a sea ice classifier based on the VGG-16 architecture and considered 5 classes (open water/leads with water, brash/pancake ice, young ice (YI), level first-year ice (FYI), and old/deformed ice). The authors have reported overall validation accuracies higher than 99%, but have not reported their test or per-class accuracy, and therefore, direct comparison is not possible. We were not able to replicate the model used in this study as the "modified VGG-16" architecture used in it was not publicly available.

[80] used a Sea Ice Residual Convolutional Network (SI-Resnet) which is based on the ResNet architecture to classify ice into four classes of (open) water, young ice, first-year ice, and old ice in Hudson Bay and the Western Arctic. Their best-performing model trained on a patch size of 57 and optimized with a multi-model average scoring strategy achieved a test accuracy of 94.05% and per-class accuracies of 97.86%, 79.41%, 92.69%, and 100%. However, they only used samples where partial concentration was above 80% and their training dataset was designed to be balanced. Our overall test accuracy is comparable with the one achieved in [80] (92% vs. 94.05%), but our per-class accuracies are lower which can be attributed to the higher number of classes in our study (6 vs 4) and the imbalance present in our training dataset. Notably, the authors in [80] state that "too fine distinction will cause much difficulty for the classification of sea ice with SAR imagery", meaning that increasing the number of classes is expected to drop performance metrics.

Even though the accuracy of our best-performing model in classifying FYI is higher than the conventional approach (0.68 vs 0.48), it is still low compared to other recent studies (for instance, in [32]). We hypothesize that this could be attributed to the low number of training samples for this class (~8% of the entire training data) as well as the inclusion of new ice and nilas classes.

Due to the small size of our public dataset, we only used 6 Sentinel-1 scenes for training our models, which is considerably lower than similar studies. For example, [36] used 299 scenes, [44] used 840 scenes, and [79] used 31 scenes. We acknowledge that this might make the model trained here not as generalizable compared to those mentioned above, but our study and results show the efficiency of our approach in training with limited samples, which can be applied to larger datasets.

Finally, it is important to note that achieving an exact, one-to-one comparison by replicating methodologies from other studies often proves challenging for multiple reasons. Firstly, many existing studies on sea ice classification utilize pre-processing and deep learning frameworks that are not open source, making it challenging to replicate methods such as those used in [79]. Secondly, the details necessary for training competing methods, such as sampling strategies and hyperparameters, often remain undisclosed in similar studies, rendering the exact replication of these methods impractical without access to such critical information. As further evidence, the authors in [29] mention that "it is difficult to compare the presented models' metric performance with other recent publications, as different training and test sets are utilized." The authors in [30] also state that "we cannot directly compare our results to other studies, which frequently use privately owned imagery and labels for training. Furthermore, the number of ice classes is different in most studies, and the polygons are drawn with different objectives and at different sizes, which make comparison difficult". Therefore, the comparisons presented in this Section should be considered with caution given these limitations.

## 4. DISCUSSION

In Section 3.1, we presented the performance of a model that uses CCE loss and one-hot encoded labels (the conventional approach in sea ice classification). Even though this model seemingly achieves a relatively high weighted average F-1 score and other aggregated performance metrics (the second row in Table II), the confusion matrix clearly shows that its performance is not consistent across different ice types, and it has a precision of below 50% in classifying both nilas and first-year ice (Fig. 7b). Our approach resulted in a higher weighted average F-1 score (and higher test accuracy and weighted average recall while keeping weighted average precision at least as high as other models), as well as a more consistent performance across different classes (with a minimum per-class precision of 60%).

There seems to be no noticeable overfitting in the models presented in Table II and Table III (when class weights are not used). The models presented in Table IV, however, generally demonstrate overfitting as there is a noticeable difference between the training and test accuracy in these models. Furthermore, our best-performing model does not demonstrate strong overfitting. Our initial experiments showed that including batch normalization in tandem with dropout in the fully connected part of the network (described in Section 2.3) leads to an average of 8% improvement in the test accuracy of the models.

We identified $\alpha = 0.25$ and $\gamma = 1$ as the optimal values for the hyperparameters of focal loss for sea ice classification. This differs from the findings of the original focal loss paper [52], in which the authors recommend $\alpha = 0.25$ and $\gamma = 2$ for general vision tasks on the COCO benchmark. We acknowledge that our findings are valid in our study design, and hyperparameters should be tuned for any specific application.

As mentioned before, we repeated each experiment twice to reduce the effect of stochasticity. Among all the experiments we performed, the maximum and average difference between testing accuracy is 4% and 2.2% respectively. For the weighted average F-1 score, the maximum and average differences are 3% and 1.8%. Weighted average precision demonstrated the lowest amount of variation with a maximum of 1% and an average of 0.5%. Finally, weighted average recall demonstrated a similar pattern to testing accuracy.

Regardless of the specific approach used for classification, the current format of ice chart polygon labels (egg code) presents a significant challenge for automated algorithms. We mentioned these challenges in Section 1. Our experiments, especially those presented in Section 3.4, provide further evidence for these challenges. Most importantly, the (usually) large spatial extent of ice chart polygons and the presence of different ice types within a polygon mean that one cannot confidently assign the label of the polygon to its constituting patches, and this would inevitably lead to label error (noise).

The models trained with partial labels and focal loss were faster by an average of 5% in terms of training time compared to models trained with CEE loss and one-hot encoded labels (21.14 seconds vs 22.2 seconds per epoch on average). Using class weights increased the training time, but not significantly (an increase of 0.2 to 0.4 seconds on average) when other variables were fixed. The choice of the loss function proved to be the most important factor in determining the convergence speed; models trained with focal loss converged anywhere from 50 to 100 epochs faster compared to those trained with CEE loss. The label encoding method made little difference in the convergence speed of the models trained with either CCE or focal loss.

We chose the ExtremeEarth v2.0 dataset for its higher spatial and thematic resolution labels, despite its fewer training data compared to other datasets like the ASIP dataset [81] or the AutoICE Challenge dataset [82].

The higher quality of the partial concentration labels in this dataset was crucial within our framework as the partial labels are generated based on partial concentration values. Furthermore, by using this dataset, we have tried to minimize the amount of inaccuracy caused by the process described above. This, however, does not completely remove such a source of uncertainty. Our method and approach are generalizable and can be applied to the datasets mentioned earlier as well, as long as partial ice concentration information is provided.

Moreover, while it is common in the literature to segment Sentinel-1 images into smaller patches, such as 50x50 patches in this study and [36], 45x45 in [40], and 32x32 in [79], this approach may result in a loss of spatial context necessary for distinguishing between wind roughening and sea ice. This justifies a future research prospect in comparing pixel-level semantic segmentation frameworks with patch-based classification frameworks in the context of sea ice mapping.

As mentioned in Section 1, different ice analysts (or the same analyst at different times) may have different interpretations of the same SAR image. This subjectivity reduces the reliability of sea ice type labels, and could potentially lead to inter-annotator disagreement, and in turn, label uncertainty. We acknowledge that the benchmark dataset used in this study does not provide inter-annotator agreement metrics, and therefore, we have not accounted for this potential source of uncertainty in our approach. Deep learning models, less sensitive to random noise than systematic biases, minimally affect results if inter-annotator disagreements are random. Ice experts typically undergo rigorous training and coordination, and typically, the work of one ice analyst is checked and confirmed by at least another analyst before making ice charts available for use.

## 5. CONCLUSIONS

In this paper, we proposed confidence-aware partial label learning as a novel approach for training deep learning classifiers in remote sensing applications where ground-truth labels are generated at the polygon level, with multiple candidate labels, each with varying levels of confidence. We tested our approach on sea ice type classification task using Sentinel-1 SAR imagery, and by performing 34 experiments, showed that: a) confidence-aware partial encoding of the labels leads to better sea ice classification performance in terms of weighted average F-1 score as well as all other aggregated performance metrics regardless of the loss function (Tables II, III, and IV), and b) integrating confidence-aware partial encoding with focal loss, when the hyperparameters of focal loss are tuned, yields better classification performance in terms of weighted average F-1 score (3% improvement), overall testing accuracy (5% improvement), and per-class F-1 score (improvement in 4 out of 6 classes) compared to the conventional approach of using one-hot encoded labels with CCE loss (Table III and Fig. 7). By allowing the training labels to be encoded as a set of candidate labels instead of one true label, our approach can take advantage of all the information embedded in labeled (annotated) polygons. By integrating partial labels with focal loss, it can (better) deal with the class imbalance which is a common issue in real-world remote sensing applications. Finally, by incorporating the confidence associated with each candidate label in encoding the partial label vector, our approach can cope with scenarios where training labels have varying levels of confidence. Our proposed approach can be beneficial in other remote sensing applications, such as land use and land cover mapping, vegetation mapping, or snow cover mapping.

Our best-performing model generates higher accuracies for FYI, old ice, open water, and nilas classes. The first three classes are usually of higher importance for operational purposes, e.g., in marine navigation, and can potentially contribute to making shipping in the Arctic safer.

The focus of this work is to show the potential of partial label encoding combined with focal loss in a PLL framework compared to the conventional one-hot encoding approach. Our high-resolution benchmark dataset contained a significantly lower amount of training data compared to similar sea ice classification studies using Sentinel-1 SAR. To address the issue with the low number of training samples, we used a pretrained ResNet-50 network. We hope that the performance gain presented in this work will encourage readers to consider the proposed framework in similar applications with more training data.

Future directions include improving melt season classification, combining partial label learning with multi-label learning in applications where each sample is associated with multiple correct labels [74], exploring the potential of data fusion, e.g. fusing C-band SAR images with other sources such as L-band (or X-band) SAR images [83], [84], and transfer learning across regions and datasets.

SUPPLEMENTARY INFORMATION

### A. Setinel-1 image information

As mentioned in Section 2.4.1, we used the 6 images acquired during the freeze-up season from ExtremeEarth dataset to generate our training, evaluation, and test sets. These images were acquired in the months of January, February, March, October, November, and December of 2018. Supplementary Table I provides the acquisition times and the file names of these images. Supplementary Fig. 1 provides an example of a Sentinel-1 SAR image acquired in Extra Wide (EW) swath mode.

Supplementary Table I

THE LIST OF THE SENTINEL-1 IMAGES USED IN EXPERIMENTS AND THEIR ACQUISITION DATES AND TIMES

| Month | Acquisition date/time | Sentinel-1 Filename |
|---|---|---|
| January | 2018-01-16 07:54:30 | S1A_EW_GRDM_1SDH_20180116T075430_20180116T075530_020177_0226B9_9FE3 |
| February | 2018-02-13 17:54:44 | S1B_EW_GRDM_1SDH_20180213T175444_20180213T175544_009608_011511_8266 |
| March | 2018-03-13 18:12:25 | S1A_EW_GRDM_1SDH_20180313T181225_20180313T181325_021000_0240E1_8163 |
| October | 2018-10-16 07:29:58 | S1A_EW_GRDM_1SDH_20181016T072958_20181016T073058_024158_02A460_DA8F |
| November | 2018-11-13 07:45:29 | S1B_EW_GRDM_1SDH_20181113T074529_20181113T074629_013583_019254_D382 |
| December | 2018-12-18 07:54:37 | S1A_EW_GRDM_1SDH_20181218T075437_20181218T075537_025077_02C472_1DB2 |

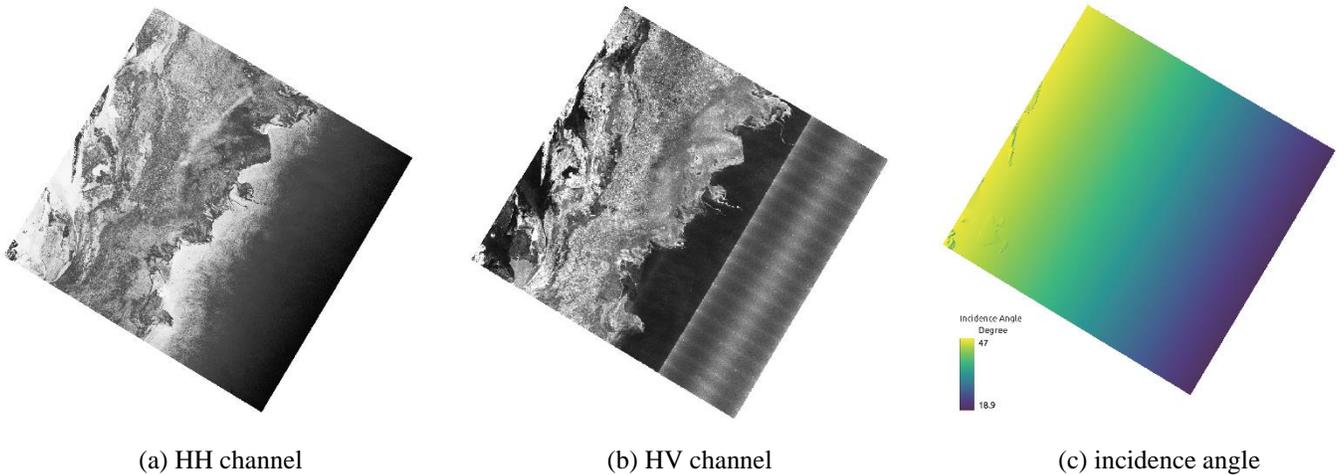

(a) HH channel     (b) HV channel     (c) incidence angle

**Supplementary Fig. 1.** An example of a Sentinel-1 image acquired in EW swath mode on January 16, 2018. (a) shows the co-polarized HH channel of the image, (b) shows the cross-polarized HV channel, and (c) shows the local incidence angle values at each pixel.

### B. Sea Ice Classification with Sentinel-1 SAR

Sea ice classification using SAR is generally performed on images acquired in C-band ($3.75 \leq \lambda \leq 7.5$ cm), and in particular, images acquired by Sentinel-1 [36], [43], [44], [79] and RADARSAT-2 instruments [20], [85]. However, ice-type separability for certain classes in this band is low, especially during the melt season [86]. L-band ($15 \leq \lambda \leq 30$ cm) and X-band frequencies ($2.5 \leq \lambda \leq 3.75$ cm) have been shown to provide useful information in those conditions, especially when complemented with C-band images [83]. However, C-band SAR remains the primary mode of acquisition in the high latitudes of the Arctic for the foreseeable and immediate future.

Sentinel-1 SAR in the Extra Wide (EW) swath mode can operate in dual-polarization mode, transmitting signals in one polarization (either horizontal or vertical) and receiving them in both the same polarization and the orthogonal polarization. This results in combinations of polarizations in the received images, such as Horizontal-Horizontal (HH) and Horizontal-Vertical (HV), or Vertical-Vertical (VV) and Vertical-Horizontal (VH).

These images, however, suffer from substantial systematic noise patterns, especially in cross-polarized images (e.g., HV polarization) over areas with low signal-to-noise ratios such as the open water or newly-formed sea ice [39], [87], [88]. Even though the cross-polarized mode is crucial in detecting certain types of ice [87], [89], the amount of noise limits the successful use of these images in machine learning-based tasks such as sea ice classification. Several approaches to denoise Sentinel-1 images have been proposed [87], [88] but, so far, these methods do not completely remove the noise and leave some residual noise. The residual noise can be misclassified as ice artifacts by automated algorithms [36]. As a result, in this work, we not only denoise the HV channel but also remove areas with high residual noise manually to minimize the effect of noise on our results (more detail in Section 3.4.1).

Sentinel-1 SAR is a side-looking imaging radar, therefore the radar waves emitted from the instruments meet the surface at a non-zero angle. The angle between the vertical direction and the radar beam direction at the surface is called the incidence angle (IA). The IA range for Sentinel-1 EW mode is 18.9° - 47.0°. It is shown that backscatter intensity decreases as IA increases, at a rate that is dependent on the type of ice [90]. Therefore, in addition to HH and HV channels, we also use local IA measurements as a predictive feature in our sea ice classification model.

Manual sea ice charting follows the "egg-code" scheme, wherein operational ice charts codify ice concentration, form, and Stage of Development (SoD) numerically into fields based on the called Sea Ice Grid 3 (SIGRID-3) format [1]. In this format, the first letter denotes the ice property (S for SoD, C for concentration, and F for floe size) and the second letter denotes the ice stage (A for primary and B for secondary). For instance, 'SA' and 'SB' represent the SoD of the oldest and second oldest ice types, respectively, while 'CA' and 'CB' denote their concentrations. The numeric codes for SA and SB represent a unique ice type (category) whereas the numeric codes used for CA and CB) represent a range of values in 10 or 20-percent increments. When following the egg code, ice analysts classify sea ice based on its SoD which is a proxy for ice thickness and age. This codification method allows for a detailed representation of ice conditions, accommodating multiple ice types within a single polygon, thus presenting challenges and opportunities for training more nuanced and accurate sea ice classification algorithms.

Supplementary Fig. 2 shows a high-resolution ice chart generated by the Norwegian MET specifically for training automated algorithms [76]. In this case, the highlighted polygon in Supplementary Fig. 2b contains two different types of ice, first-year ice, and young ice, with partial concentrations ranging between 70–90% and 20–40% respectively.

Labeling Sentinel-1 images at the pixel level can be exponentially more expensive (potentially even prohibitively so). Each Sentinel-1 SAR image in Extra-Wide swath mode contains approximately 10,000×10,000 pixels, with a resolution of 93×87 m along the range and azimuth directions (and a pixel spacing of 40×40 m). As such, it would present a very challenging and time-consuming endeavor to assign labels at the pixel level, even for a trained expert.

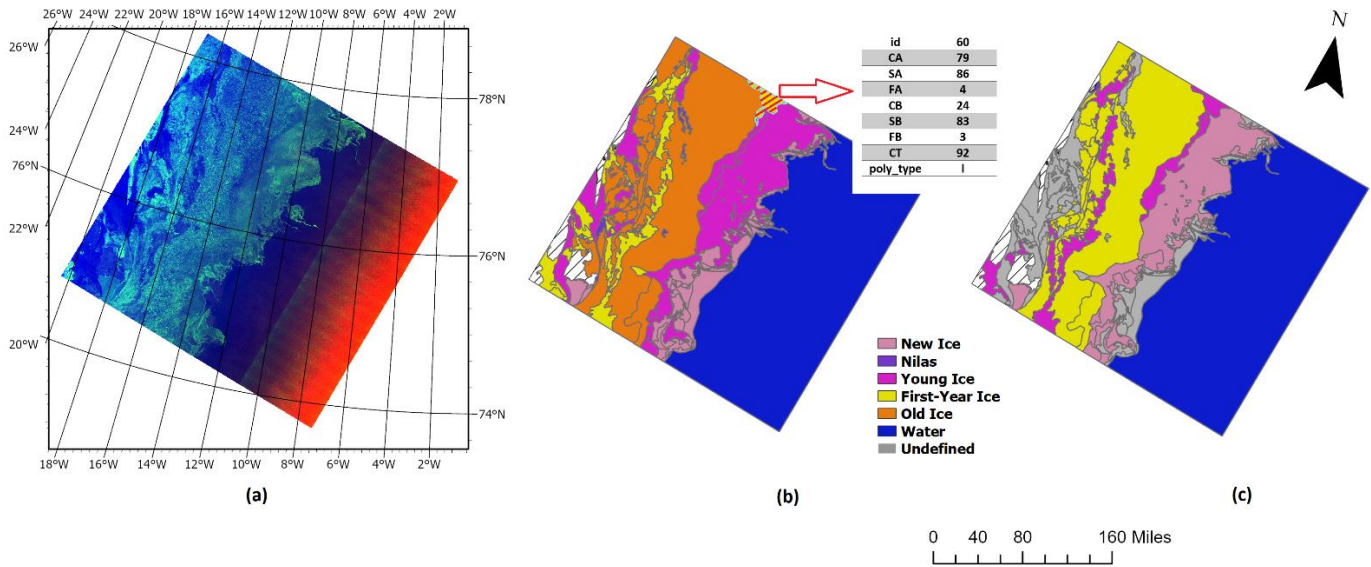

**Supplementary Fig. 2.** (a) A false-color example of a Sentinel-1 SAR image acquired on January 16, 2018. R: HH co-polarized channel, G: HV cross-polarized channel, B: local incidence angle, c) the corresponding ice chart from ExtremeEarth dataset [11]. CA and CB represent partial sea ice concentration of the primary (oldest) and secondary (youngest) ice types respectively. SA and SB represent the Stage of Development (SoD) of those two ice types, and FA and FB represent the form of ice (or floe size). CT represents total concentration (CA+CB). In (b) polygons are color-coded based on the oldest SoD present in them and in (c) based on the second oldest SoD. The numeric values represent a range in the case of concentrations and a type in the case of SA/SB (please refer to Section 2.4.1 for more information).

*C. The Effect of FL hyperparameters on model performance*

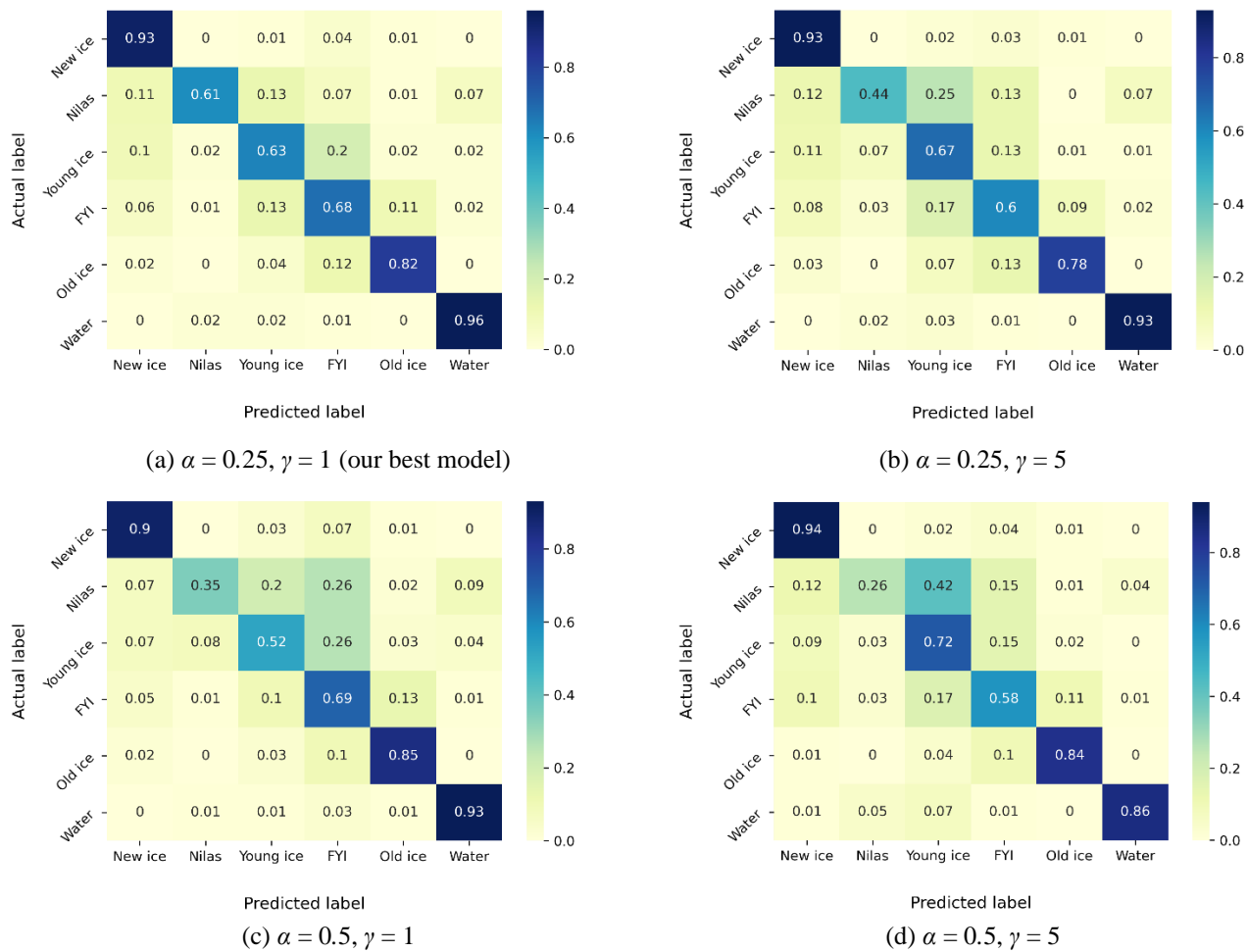

**Supplementary Fig. 3.** Confusion Matrices for 4 of the models trained with confidence-aware partial labels and focal loss (α ∈ {0.25, 0.5} and γ ∈ {1, 5}).

Fig. 7a presented the confusion matrix of our best-performing model in the group of models trained with confidence-aware partial labels and foal loss (the third row in Table I). To examine the influence of focal loss hyperparameters on the per-class performance of this group of models, Supplementary Fig. 3 contrasts the confusion matrices of this model with three other models from Table III that have α ∈ {0.25, 0.5} and γ ∈ {1, 5} (the same models that are utilized in Fig. 6). The observations indicate that modifying α or γ may enhance performance for specific classes; however, the consistency of models in b-d is inferior to model a, emphasizing the overall advantage of our selected configuration (α = 0.25, γ = 1). This further underscores the necessity of fine-tuning FL hyperparameters based on the dataset, and when potentially prioritizing a particular ice type over others.

*D. Focal Loss Sensitivity Analysis*

Section 3.3 presented the discussion on the sensitivity of FL to α and γ hyperparameters. Supplementary Fig. 4 presents the training loss and accuracy plots for the set of hyperparameters used in our experiment.

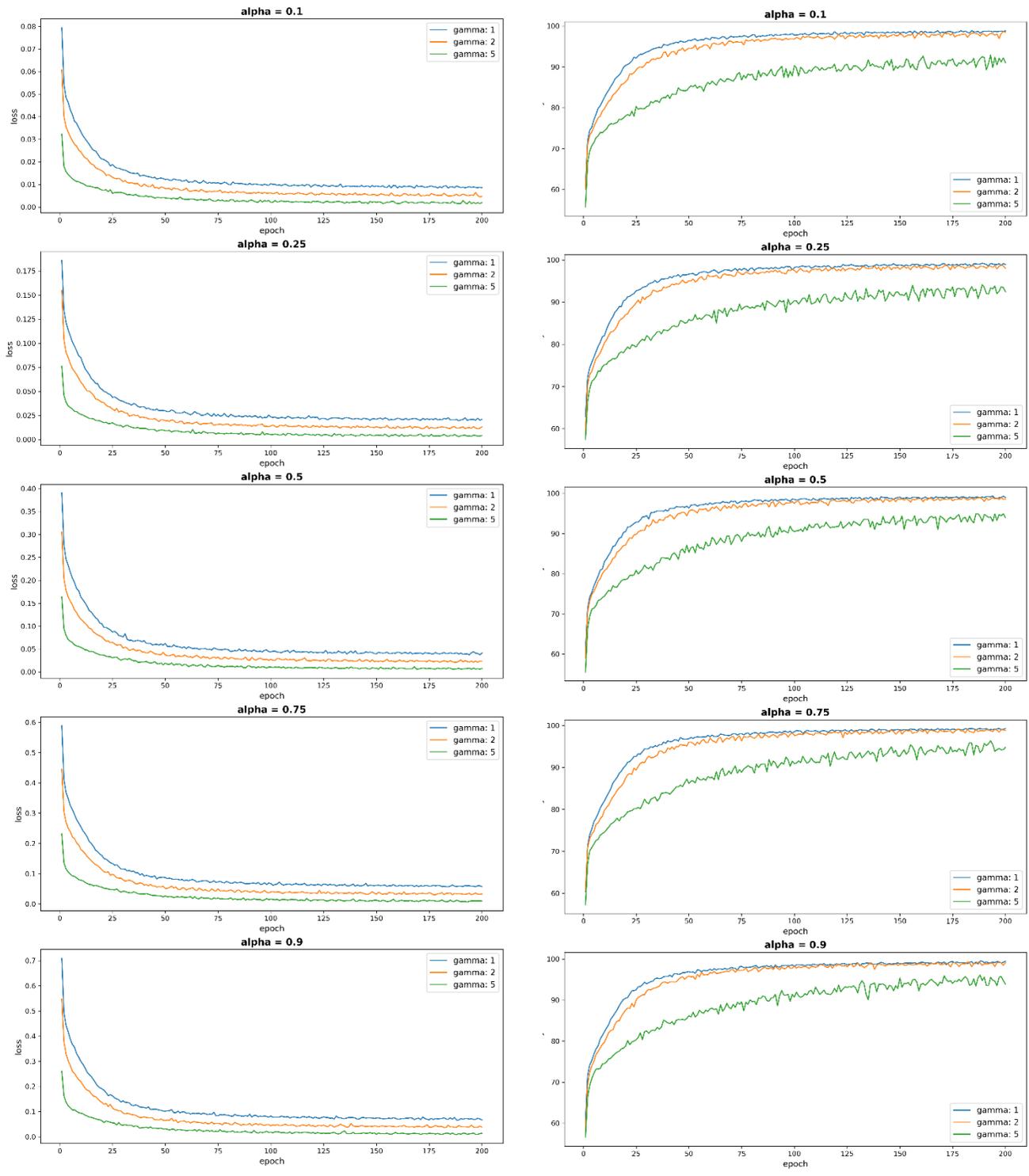

(a) Loss                                                    (b) Accuracy

**Supplementary Fig. 4.** Sensitivity of training loss and accuracy to the hyperparameters of focal loss (α and γ).